\pdfoutput=1

\documentclass[11pt]{article}

\usepackage[final]{acl}

\usepackage{times}
\usepackage{latexsym}

\usepackage[T1]{fontenc}

\usepackage[utf8]{inputenc}

\usepackage{microtype}

\usepackage{inconsolata}

\usepackage{graphicx}
\usepackage{multirow}

\usepackage{amsmath}
\usepackage{amsfonts}
\usepackage{bbm}
\usepackage{booktabs}
\usepackage{algorithm}
\usepackage{algpseudocode}
\usepackage{hyperref}
\usepackage{cleveref}
\usepackage[most]{tcolorbox}

\usepackage{tabularx} 

\title{Mirror-Consistency: Harnessing Inconsistency in Majority Voting}

  \author{
      Siyuan Huang\textsuperscript{1,3,\footnotemark[1]},
      Zhiyuan Ma\textsuperscript{2},
      Jintao Du\textsuperscript{2}\\
      \textbf{
      Changhua Meng\textsuperscript{2},
      Weiqiang Wang\textsuperscript{2},
          Zhouhan Lin \textsuperscript{1,\footnotemark[2]}} \\
    \textsuperscript{\rm 1} Shanghai Jiaotong University\\
    \textsuperscript{\rm 2} Tiansuan Lab, Ant Group Co., Ltd.\\
     \textsuperscript{\rm 3}  SJTU Paris Elite Institute of Technology \\
\texttt{siyuan\_huang\_sjtu@outlook.com}, \texttt{lin.zhouhan@gmail.com}\\ \texttt{\{mazhiyuan.mzy,lingke.djt,changhua.mch,weiqiang.wwq\}@antgroup.com}\\}

\begin{document}
\maketitle

\renewcommand{\thefootnote}{\fnsymbol{footnote}} 
\footnotetext[1]{Work done during an internship at Ant Group.}
\footnotetext[2]{Corresponding Author.}

\begin{abstract}
Self-Consistency, a widely-used decoding strategy, significantly boosts the reasoning capabilities of Large Language Models (LLMs). However, it depends on the plurality voting rule, which focuses on the most frequent answer while overlooking all other minority responses. These inconsistent minority views often illuminate areas of uncertainty within the model's generation process. To address this limitation, we present Mirror-Consistency, an enhancement of the standard Self-Consistency approach. Our method incorporates a `reflective mirror' into the self-ensemble decoding process and enables LLMs to critically examine inconsistencies among multiple generations. Additionally, just as humans use the mirror to better understand themselves, we propose using Mirror-Consistency to enhance the sample-based confidence calibration methods, which helps to mitigate issues of overconfidence. Our experimental results demonstrate that Mirror-Consistency yields superior performance in both reasoning accuracy and confidence calibration compared to Self-Consistency.
\end{abstract}

\section{Introduction}

\begin{figure}[t]
  \includegraphics[width=\columnwidth]{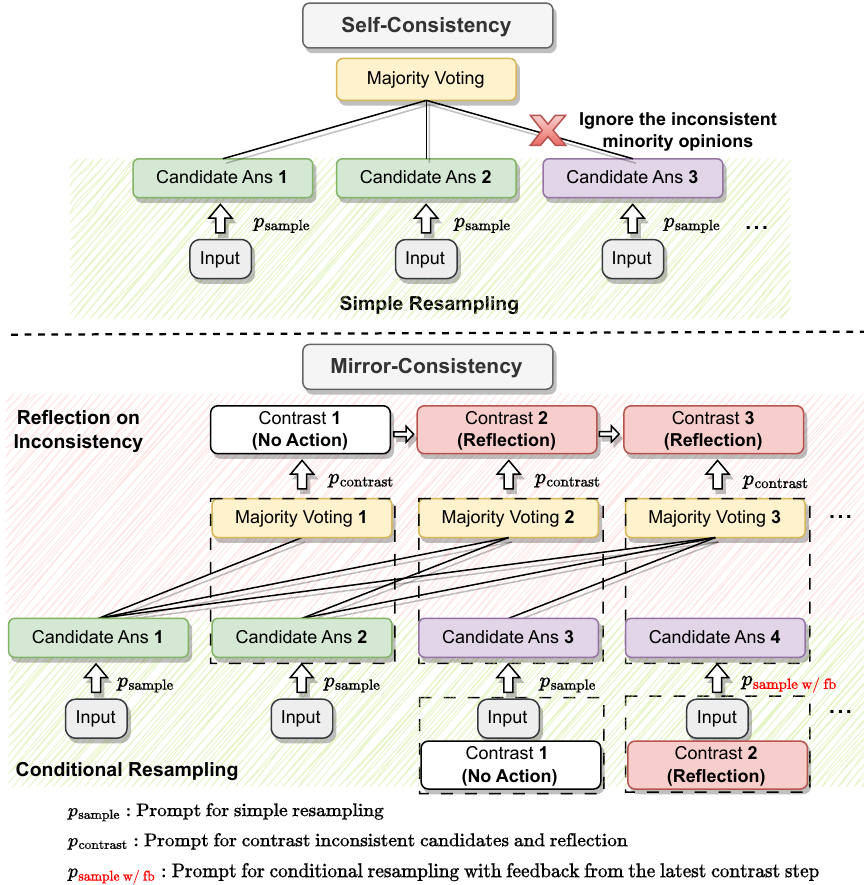}
  \caption{Mirror-Consistency vs. Self-Consistency. While Self-Consistency overlooks minority responses, Mirror-Consistency takes them into account by reflecting on inconsistencies  during the sampling process.}
  \label{fig:mirror-consistency}
\end{figure}

Large Language Models (LLMs) have substantially influenced diverse sectors with their advanced applications~\citep{chowdhery2022palm, schick2023toolformer, wu2023visual, shen2023hugginggpt, Zhang2023DataCopilotBB}. To further bolster LLMs' reasoning ability, Self-Consistency~\citep{wang2023selfconsistency} initially generates a wide range of reasoning pathways, then performs a marginalization to determine the most consistent response. Although generally effective, this approach relies primarily on the plurality voting rule, which focuses only on the most frequent answer, thereby neglecting other minority responses. Consequently, the crucial inconsistencies in the resampled answers, which could reveal uncertainties and potential errors of LLMs, often do not receive the attention they merit.

To address this limitation, we introduce Mirror-Consistency, which enables the LLM to analyze and learn from discrepancies in its resampled responses. The complete process is illustrated in \autoref{fig:mirror-consistency}. We sequentially resample new responses. Each new response is then compared to the previous majority voting result. If the new response deviates from the majority's decision (i.e., it represents a minority opinion), we position the LLM to act as an arbitrator. In this role, the LLM reflects on the differences between the new response and the previous majority opinion and offers a feedback, which is then utilized to guide the generation of subsequent responses. This process effectively equips the standard Self-Consistency method with a `mirror', enabling the LLM to learn from inconsistencies observed during the sampling process. 

We then delve into another application of Mirror-Consistency: the confidence calibration of LLMs~\citep{desai-durrett-2020-calibration, geng_survey_2023}. Recent studies leverage the consistency of multiple generations to assess the LLM's confidence in its responses~\citep{DBLP:journals/corr/abs-2306-13063, wang2023selfconsistency, selfcheckgpt, DBLP:journals/corr/abs-2306-13063, portillo-wightman-etal-2023-strength}. Following this idea, we find that Mirror-Consistency, which involves repeatedly comparing and reflecting on the discrepancies between the majority and the minority opinions, provides a more robust quantification of uncertainty, particularly in cases of overconfidence. Our motivation for using mirror consistency in confidence calibration stems from the idea that answers that consistently emerge as the majority after several rounds of comparisons with other candidate answers are likely more reliable.

We conduct experiments on four reasoning datasets with four different LLMs to compare the performance of Mirror-Consistency with Self-Consistency. The results confirm the efficacy of reflecting on the inconsistencies during the sampling process, showing improvements in both reasoning accuracy and confidence calibration.

\section{Mirror-Consistency}

\begin{table*}[!t]
\footnotesize 
\centering
\renewcommand{\arraystretch}{0.9} 
\scalebox{0.96}{ 
\begin{tabular}{@{}l||l||cc||cccc@{}} 
\toprule
Model           & Method                               & \# Ans & \# Call & GSM8K & SVAMP & StrategyQA & Date \\ \midrule
\textbf{GPT-3.5-turbo} & Standard COT                         & 1 & 1 & 75.9  & 77.3  & 73.6       & 67.3 \\
                      & Self-Consistency                     & 10 & 10 & 85.2~\tiny{$\uparrow$9.3}  & 84.2~\tiny{$\uparrow$6.9}  & 77.9~\tiny{$\uparrow$4.3}       & 72.4~\tiny{$\uparrow$5.1} \\
                      & Self-Consistency                     & 20 & 20 & 85.4~\tiny{$\uparrow$9.5}  & 84.5~\tiny{$\uparrow$7.2}  & 77.8~\tiny{$\uparrow$4.2}        & \textbf{73.0~\tiny{$\uparrow$5.7}} \\
\cmidrule{2-8}
                      & Mirror-Consistency                   & 10 & 19 & \textbf{86.7~\tiny{$\uparrow$10.8}}  & \textbf{86.1~\tiny{$\uparrow$8.8}}  & \textbf{78.4~\tiny{$\uparrow$4.8}}       & 72.6~\tiny{$\uparrow$5.3}  \\
\midrule
\textbf{Qwen-turbo}   & Standard COT                         & 1 & 1 & 74.5  & 76.9  & 71.9       & 62.7 \\
                      & Self-Consistency                     & 10 & 10 & 77.4~\tiny{$\uparrow$2.9}  & 79.1~\tiny{$\uparrow$2.2}  & 73.8~\tiny{$\uparrow$1.9}       & 63.5~\tiny{$\uparrow$0.8} \\
                      & Self-Consistency                     & 20 & 20 & 77.9~\tiny{$\uparrow$3.4}  & 79.1~\tiny{$\uparrow$2.2}  & 73.6~\tiny{$\uparrow$1.7}       & 63.5~\tiny{$\uparrow$0.8} \\
\cmidrule{2-8}
                      & Mirror-Consistency                   & 10 & 19 & \textbf{79.7~\tiny{$\uparrow$5.2}}  & \textbf{83.0~\tiny{$\uparrow$6.1}}  & \textbf{74.7~\tiny{$\uparrow$2.8}}       & \textbf{64.3~\tiny{$\uparrow$1.6}} \\
\midrule
\textbf{Llama3-8B}   & Standard COT                         & 1      & 1       & 78.7  & 80.8  & 71.6       & 76.9 \\
                      & Self-Consistency                     & 10     & 10      & 85.9~\tiny{$\uparrow$7.2}  & 89.0~\tiny{$\uparrow$8.2}  & 72.0~\tiny{$\uparrow$0.4}       & 83.8~\tiny{$\uparrow$6.9} \\
                      & Self-Consistency                     & 20     & 20      & 86.5~\tiny{$\uparrow$7.8}  & 89.6~\tiny{$\uparrow$8.8}  & \underline{72.8~\tiny{$\uparrow$1.2}} & \textbf{84.1~\tiny{$\uparrow$7.2}} \\
\cmidrule{2-8}
                      & Mirror-Consistency                   & 10     & 19      & \textbf{87.2~\tiny{$\uparrow$8.5}} & \textbf{90.4~\tiny{$\uparrow$9.6}}  & \underline{72.8~\tiny{$\uparrow$1.2}} & 83.5~\tiny{$\uparrow$6.6}  \\
\midrule
\textbf{Llama3-70B}  & Standard COT                         & 1      & 1       & 92.7  & 91.8  & 72.9       & 90.6 \\
                      & Self-Consistency                     & 10     & 10      & 95.3~\tiny{$\uparrow$2.6}  & 93.5~\tiny{$\uparrow$1.7}     & 73.7~\tiny{$\uparrow$0.8}          & \underline{94.2~\tiny{$\uparrow$3.6}}    \\
                      & Self-Consistency                     & 20     & 20      & 95.6~\tiny{$\uparrow$2.9}  & \textbf{94.0~\tiny{$\uparrow$2.2}}     & 73.7~\tiny{$\uparrow$0.8}          & \underline{94.2~\tiny{$\uparrow$3.6}}    \\
\cmidrule{2-8}
                      & Mirror-Consistency                   & 10     & 19      & \textbf{95.8~\tiny{$\uparrow$3.1}}  & 93.2~\tiny{$\uparrow$1.4} & \textbf{74.7~\tiny{$\uparrow$1.8}} & \underline{94.2~\tiny{$\uparrow$3.6}} \\
\bottomrule
\end{tabular}
}
\caption{Performance comparison between Mirror-Consistency and Self-Consistency on Arithmetic Reasoning and Commensense Reasoning tasks. The best scores are bolded. Multiple identical scores are indicated with underlining.}
\label{tab:Performance comparison between Mirror-Consistency and Self-Consistency}
\end{table*}


This section details the implementation of Mirror-Consistency. Self-Consistency depends on the plurality voting rule, which tends to overlook minority responses that may be inconsistent yet informative. To address this gap, we introduce a reflective form of consistency, termed Mirror-Consistency. 

Mirror-Consistency and Self-Consistency share procedural similarities, as both methods generate multiple responses and use majority voting to determine the final output. The key distinction is that Mirror-Consistency examines the inconsistencies within the sampling process, rather than merely repeat sampling independent answers. The Mirror-Consistency process begins with an initial sampling phase to produce the first response, denoted as $r_0$. This is followed by a series of iterations that alternate between Reflection on Inconsistency and Conditional Resampling, as shown in \Cref{alg:mirror-consistency-algo}. We next introduce these two steps in detail. The corresponding prompts are provided in \autoref{sec:prompt}. 
\paragraph{Reflection on Inconsistency} During the $k^{th}$ Reflection on Inconsistency, the model compares the newly resampled response $r_k$ with the majority voting result $m_{k-1}$ from previous rounds. If the responses align, the process moves to the next resampling phase. Otherwise, the model analyzes the causes of the discrepancies and formulates a suggestion for additional checks, as the inconsistencies of multiple reasoning pathways often reveal uncertainty and potential mistakes within the generation. This suggestion is incorporated into the existing checklist $C_{k-1}$, updating it to $C_k$ for use in the subsequent Conditional Resampling.
\paragraph{Conditional Resampling}
In this phase, the feedback obtained from the previous round is utilized to guide the generation of subsequent responses. During the \(k^{th}\) round of Conditional Resampling, the checklist \(C_k\), derived from earlier rounds, is integrated into the prompt, facilitating the generation of a new response \(r_{k+1}\). This method ensures that the LLM concentrates on areas of uncertainty exposed by inconsistencies noted during prior samplings.

\begin{algorithm}[H] 
\caption{Mirror-Consistency}
\label{alg:mirror-consistency-algo}
\begin{algorithmic}[1]
\Require  prompts \{$p_{\text{sample}}, p_{\text{sample w/ fb}}, p_{\text{contrast}}$\}, model $\mathcal{M}$, input $x$
\State $r_0 \leftarrow \mathcal{M}(p_{\text{sample}}\|x)$ \Comment{Initial response}
\State $C_0 \leftarrow \emptyset$ \Comment{Initial checklist}
\State $m_0 \leftarrow r_0$ \Comment{Initial majority vote}
\For{$k \in \{1, \dots, K\}$}
    \State $r_k \leftarrow \mathcal{M}(p_{\text{sample w/ fb}}\|C_{k-1}\|x)$ 
    \Statex \Comment{Conditional Resampling}
    \State $C_k \leftarrow \mathcal{M}(p_{\text{contrast}}\|m_{k-1}\|r_{k}\|C_{k-1}\|x)$ 
    \Statex \Comment{Reflection on Inconsistency}
    \State $m_k \leftarrow \text{majority-vote}(r_0, \ldots, r_k)$ 
    \Statex \Comment{Update majority vote}
\EndFor
\State \Return $m_K$
\end{algorithmic}
\end{algorithm}

\paragraph{Comparison with Self-Reflection} Standard intrinsic self-correction methods directly generate feedback for the initial responses, which frequently exhibit overconfidence and difficulty in identifying errors~\citep{DBLP:journals/corr/abs-2310-01798, DBLP:journals/corr/abs-2311-08516}. In contrast, Mirror-Consistency identifies potential errors by critically analyzing the inconsistencies across multiple reasoning pathways. This contrastive strategy is generally more practical than directly generating feedback~\citep{DBLP:journals/corr/abs-2401-02009}.

\section{Calibration with Mirror-Consistency}\label{sec: Calibration with Mirror-Consistency}
In this section, we introduce the application of Mirror-Consistency to the calibration of LLMs. 
Recent studies have shown that consistency across multiple model generations serves as a reliable indicator of confidence~\citep{portillo-wightman-etal-2023-strength, selfcheckgpt}. While effective, these methods rely on the Self-Consistency logic  (i.e., `independent' repeated sampling) and overlook the analysis of inconsistencies among responses. In contrast, Mirror-Consistency employs a \emph{reflective form of consistency}: responses maintaining consistency after numerous comparisons and reflections are deemed more reliable; while responses that vary upon comparison and reflection highlight the LLM’s inherent uncertainties. Hence, we propose using the distribution of responses generated by Mirror-Consistency as a measure of confidence.

Next, we introduce two fundamental types of sample-based confidence metrics. A comprehensive analysis of five existing sample-based confidence metrics is provided in~\autoref{sec: Additional Calibration Results}. The most commonly employed strategy is the agreement-based metric~\citep{DBLP:journals/corr/abs-2402-13904}. For each input \(x\), we generate \(n\) candidate outputs, \(\hat{r}_1, \dots, \hat{r}_n\). We then apply majority voting across the answers to determine the most-voted answer \(\bar{r} = \text{argmax}_r \sum_{i=1}^n \mathbbm{1} (\hat{r}_i = r)\). The agreement-based confidence score is then defined as the percentage of answers that agree with the most-voted answer:
\[
   \texttt{Agree}(\bar{r}) =  \frac{1}{n}\sum_{i=1}^n \mathbbm{1} (\hat{r}_i = \bar{r})
\]
$\texttt{Agree}$ solely relies on the most voted answer. \textbf{F}irst-\textbf{S}econd-\textbf{D}istance (FSD)~\citep{DBLP:journals/corr/abs-2402-13904} provides another option by considering the top two most-voted answers, denoted as \(\bar{r}_1\) and \(\bar{r}_2\). \(\texttt{FSD}(\bar{r})\) is the difference in their agreement rates:
\[
\texttt{FSD}(\bar{r}) = \texttt{Agree}(\bar{r}_1) - \texttt{Agree}(\bar{r}_2)
\]
Further experiments, as detailed in \autoref{sec:calibration} and \autoref{sec: Additional Calibration Results}, demonstrate that this reflective form of consistency provides more reliable calibration than standard sample-based methods.

\section{Experiments}


\begin{figure*}[t]
  \includegraphics[width=\linewidth]{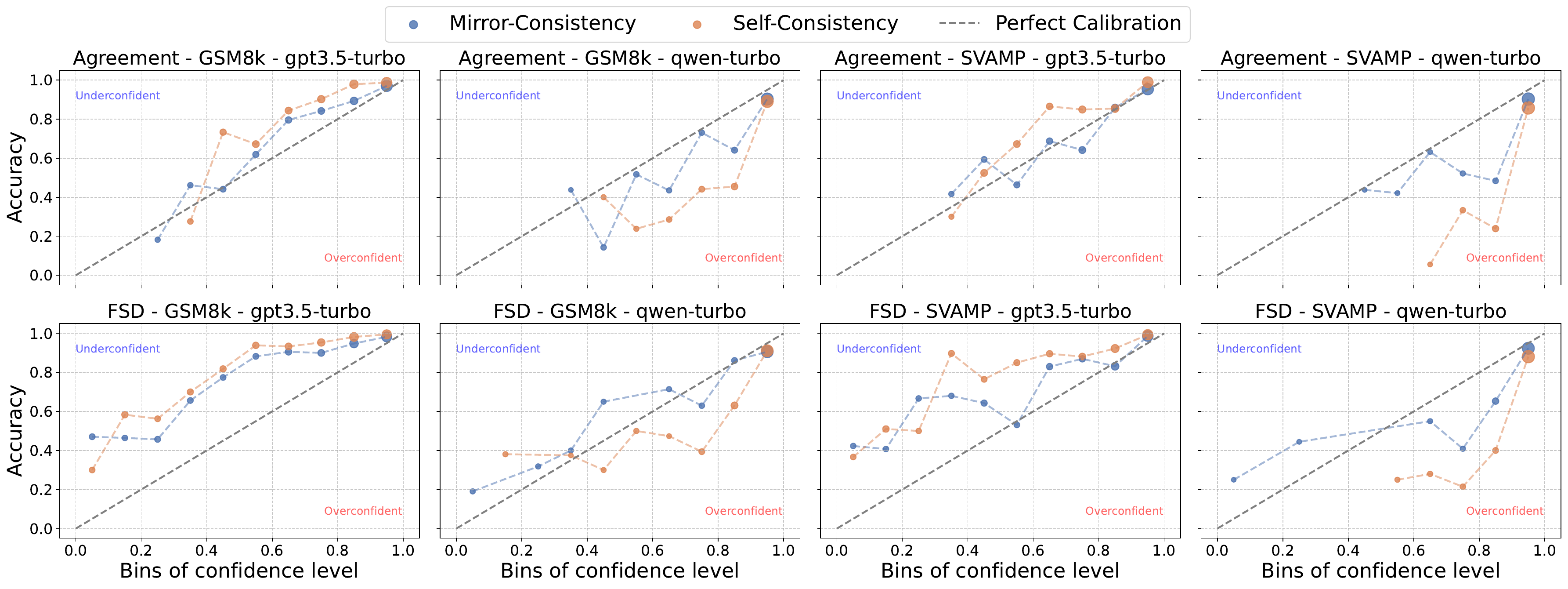}
  \centering
\caption{Comparison of calibration curves on GSM8K and SVAMP with \texttt{Agree} and \texttt{FSD}. We omit the bins if their size is less than 1\% of the total dataset. The point size is proportional to the square root of its corresponding bin size. }
\label{fig:ECE Visualization}
\end{figure*}

In this section, we assess the efficacy of Mirror-Consistency from two distinct angles: First, we evaluate its reasoning accuracy in comparison to Self-Consistency across four reasoning datasets. Second, we demonstrate how the reflective form of consistency embodied by Mirror-Consistency improves the sample-based calibration methods.

\subsection{Reasoning Accuracy}\label{sec: Reasoning Accuracy}

\paragraph{Benchmarks} We evaluate our models using four distinct datasets: GSM8K \citep{cobbe_training_2021} and SVAMP \cite{patel_are_2021}, which focus on arithmetic reasoning; StrategyQA~\citep{geva_did_2021} and Date Understanding~\citep{big-bench_collaboration_beyond_2021}, which belong to Multi-hop QA. We follow the same split setting as \citet{fcot}.

\paragraph{LLMs} We utilize four different large language models: GPT-3.5-turbo-0613, Qwen-turbo, and the newly released Llama-3 family (8B/70B). More details can be found in \autoref{sec: Model Details}.

\paragraph{Baselines} We compare our method with Standard CoT Prompt~\citep{kojima2022large} and Self-Consistency~\citep{Wang2023Self}. To ensure fairness in our comparison, we evaluate two scenarios: equivalent number of responses (10 vs. 10), and comparable number of calls (19 vs. 20).

\paragraph{Prompting and Sampling Strategies} Throughout our experiments, we consistently use the same COT prompt, which is detailed in \autoref{sec:prompt} with a temperature $T=0.4$, following \citet{DBLP:journals/corr/abs-2402-13904}.

\paragraph{Main Result} The main results are presented in \autoref{tab:Performance comparison between Mirror-Consistency and Self-Consistency}.  Mirror-Consistency outperforms Self-Consistency, even with the latter doubling its number of responses. We observe that Mirror-Consistency generally yields greater improvements in arithmetic reasoning datasets. This can be attributed to the fact that, whereas errors in common sense are more likely constrained by the inherent limitations of LLMs, arithmetic errors are more amenable to correction through reflective methods.

\subsection{Confidence Calibration}\label{sec:calibration}

\paragraph{Experiment Setup} We compare the performance of Mirror-Consistency and Self-Consistency, each using a comparable number of calls (19 vs. 20). We consider two LMs, GPT-3.5-turbo and Qwen-turbo, on two tasks, GSM8K and SVAMP, with the same settings in \autoref{sec: Reasoning Accuracy}. We use Expected Calibration Error (ECE)~\citep{calibration-guo} as the evaluation metric. We first consider two confidence metrics, \texttt{Agree} and \texttt{FSD}, as introduced in \autoref{sec: Calibration with Mirror-Consistency}.

\paragraph{Main Result} \autoref{tab:ECE Scores Comparison} presents the ECE comparisons across two different confidence metrics. Mirror-Consistency consistently yields superior results relative to Self-Consistency. This finding suggests that reflecting on inconsistencies during the sampling process leads to more accurate confidence estimations, offering a novel approach to calibration tasks. \autoref{fig:ECE Visualization} presents the corresponding calibration curves. We find that Mirror-Consistency significantly mitigates the overconfidence issues encountered with Self-Consistency. For instance, when using responses generated by Self-Consistency, the curve of Qwen-turbo tends to the bottom right. In contrast, the curve of Mirror-Consistency is clearly closer to the ideal line.

\paragraph{Additional Results} We consider five different sample-based calibration metrics in \autoref{sec: Additional Calibration Results}. \autoref{tab: Full ECE Scores Comparison} presents a complete comparison. The corresponding calibration curves are shown in \autoref{fig: full calibration curves}.

\begin{table}
\centering
\small
\begin{tabular}{@{}l||cc|cc@{}}
\toprule
& \multicolumn{2}{c|}{Agreement} & \multicolumn{2}{c}{FSD} \\
\cmidrule(l){2-5} 
Model \& Method & \textbf{G} & \textbf{S} & \textbf{G} & \textbf{S} \\ 
\midrule
\textbf{GPT-3.5-turbo} & & & & \\ 
\quad Self-Consistency & 0.081 & 0.044 & 0.170 & 0.127 \\
\quad Mirror-Consistency & \textbf{0.039} & \textbf{0.042} & \textbf{0.138} & \textbf{0.082} \\ 
\midrule
\textbf{Qwen-turbo} & & & & \\
\quad Self-Consistency & 0.154 & 0.196 & 0.127 & 0.173 \\
\quad Mirror-Consistency & \textbf{0.102} & \textbf{0.103} & \textbf{0.089} & \textbf{0.100} \\
\bottomrule
\end{tabular}
\caption{ECE ↓ results of Mirror-Consistency and Self-Consistency. \textbf{G} denotes GSM8K and \textbf{S} denotes SVAMP.}
\label{tab:ECE Scores Comparison}
\end{table}


\section{Conclusion}
Mirror-Consistency effectively remedies a crucial limitation in the standard Self-Consistency method: the disregard for minority viewpoints during sampling. Our approach examines the discrepancies across various responses to identify uncertainties within the generative process. The experimental results demonstrates that Mirror-Consistency not only achieves higher reasoning accuracy compared to Self-Consistency but also enhances model calibration, especially when standard sample-based calibration methods face issues of overconfidence.
\section{Limitations}
We acknowledge several limitations that invite further investigation: Firstly, while Self-Consistency can generate multiple answers in parallel, Mirror-Consistency, due to the fact that each generation of a new answer needs to be compared with the previous majority opinion, must generate answers sequentially. Secondly, our experimental setup solely utilizes Chain-of-Thought~\cite{cot} as the prompt strategy for sampling. Exploring a broader array of prompt strategies could provide deeper insights into their interactions with the Mirror-Consistency approach. Thirdly, the central motivation behind Mirror-Consistency is to effectively harness the inconsistencies that emerge during repeated sampling processes. This requires the responses to exhibit enough diversity. To enhance this diversity, we could enable the large language model to automatically generate prompts that vary across different dimensions~\cite{DBLP:journals/corr/abs-2401-02009}. Fourthly, we fix the temperature parameter throughout our study. Future work could experiment with varying the temperature to assess the robustness of Mirror-Consistency under different settings. Lastly, since Mirror-Consistency relies on prompting to encourage reflective thinking, it necessitates strong instruction-following capabilities within the model. While the four models considered in our experiments demonstrate adequate ability in this regard, this strategy may not be as effective with smaller-scale or lower-performing models.

\section{Ethical Considerations}
We recognize several ethical Considerations. Our approach, Mirror-Consistency, improves LLM robustness by addressing inconsistencies but cannot fully negate the risks of errors or biases from the underlying data and model structures. Additionally, the method's iterative process, involving repeated resampling and reflection, presents concerns about computational costs and environmental impacts. Finally, it's crucial to clearly convey the limitations of Mirror-Consistency, informing users about possible failure scenarios and the potential repercussions. This transparency is vital for setting appropriate expectations for real-world applications.

\section*{Acknowledgement}
This research work has been sponsored by Ant Group Security and Risk Management Fund, the National Key Research and Development Program of China (No. 2023ZD0121402) and National Natural Science Foundation of China (NSFC) grant (No.62106143).

\bibliography{custom}

\appendix
\section{Additional Calibration Results}\label{sec: Additional Calibration Results}
\begin{table*}[!t]
\footnotesize 
\centering
\small
\begin{tabular}{@{}l||cc|cc|cc|cc|cc|c@{}}
\toprule
& \multicolumn{2}{c|}{Agreement} & \multicolumn{2}{c|}{Entropy} & \multicolumn{2}{c|}{FSD} & \multicolumn{2}{c|}{Ans-Num} & \multicolumn{2}{c|}{Pairwise} & \multicolumn{1}{c}{\# Win} \\
\cmidrule(l){2-12} 
Model \& Method & G & S & G & S & G & S & G & S & G & S & \\ 
\midrule
\textbf{GPT-3.5-turbo} & & & & & & & & & & & \\ 
\quad Self-Consistency & 0.081 & 0.044 & 0.291 & 0.223 & 0.170 & 0.127 & \textbf{0.056} & \textbf{0.049} & 0.090 & 0.057 & 2 \\
\quad Mirror-Consistency & \textbf{0.039} & \textbf{0.042} & \textbf{0.279} & \textbf{0.196} & \textbf{0.138} & \textbf{0.082} & 0.078 & 0.068 & \textbf{0.053} & \textbf{0.051} & 8 \\ 
\midrule
\textbf{Qwen-turbo} & & & & & & & & & & & \\
\quad Self-Consistency & 0.154 & 0.196 & \textbf{0.104} & 0.137 & 0.127 & 0.173 & 0.145 & 0.163 & \textbf{0.099} & 0.194 & 2 \\
\quad Mirror-Consistency & \textbf{0.102} & \textbf{0.103} & 0.138 & \textbf{0.104} & \textbf{0.089} & \textbf{0.100} & \textbf{0.039} & \textbf{0.057} & 0.150 & \textbf{0.101} & 8 \\
\bottomrule
\end{tabular}
\caption{Summary of ECE ↓ results across different calibration metrics. \textbf{G} denotes GSM8K and \textbf{S} denotes SVAMP. We show the statistics of the comparison in the last column.}
\label{tab: Full ECE Scores Comparison}
\end{table*}
In this section, we explore the impact of various metrics on calibration results based on Mirror-Consistency and Self-Consistency approaches. Recently, numerous studies have adopted sample-based methods to assess the confidence of Large Language Models (LLMs)~\citep{wang2023selfconsistency, selfcheckgpt, DBLP:journals/corr/abs-2306-13063, portillo-wightman-etal-2023-strength}. These studies have proposed different metrics to derive corresponding confidence scores from the varying resampled responses. Again, for each input \(x\), we generate \(n\) candidate outputs, \(\hat{r}_1, \dots, \hat{r}_n\). We then apply majority voting across the answers to determine the most-voted answer \(\bar{r} = \text{argmax}_r \sum_{i=1}^n \mathbbm{1} (\hat{r}_i = r)\). We consider the following five metrics:\\
\(\bullet\) \emph{Agreement-based}~\citep{DBLP:journals/corr/abs-2402-13904}:\\
The agreement-based confidence score, which is the most common metric and has been introduced in \autoref{sec: Calibration with Mirror-Consistency}, is defined as the percentage of answers that agree with the most-voted answer:
\[
   \texttt{Agree}(\bar{r}) =  \frac{1}{n}\sum_{i=1}^n \mathbbm{1} (\hat{r}_i = \bar{r})
\]
\(\bullet\) \emph{Entropy-based}~\citep{DBLP:journals/corr/abs-2402-13904}: \\
We first derive a set of unique answers from the model’s output, denoted as \(\mathbf{\hat{r}}\), by eliminating duplicates. The entropy-based consistency, \(\texttt{Ent}(\bar{r})\), is then defined as follows:
\[
\texttt{Ent}(\bar{r}) = 1 - \left(-\frac{1}{\log |\mathbf{\hat{r}}|} \sum_{i=1}^{|\mathbf{\hat{r}}|} p_i \log p_i\right)
\]
Here, \(|\mathbf{\hat{r}}|\) represents the number of unique answers, and \(p_i\) is the normalized frequency of each unique answer \(\hat{r}_i\) in the dataset. This formulation inversely relates the entropy measure to the consistency of the answer distribution, where a lower entropy implies a more uniform and certain response pattern.\\
\(\bullet\) \emph{FSD-based}~\citep{DBLP:journals/corr/abs-2402-13904}: \\
FSD stands for \textbf{F}irst-\textbf{S}econd-\textbf{D}istance. To calculate the FSD-based consistency, we first identify the top two most-voted answers, denoted as \(\bar{r}_1\) and \(\bar{r}_2\). We then calculate the agreement rates for these two answers, denoted as \(\texttt{Agree}(\bar{r}_1)\) and \(\texttt{Agree}(\bar{r}_2)\), respectively. The FSD-based consistency, \(\texttt{FSD}(a)\), is computed as the difference in their agreement rates:
\[
\texttt{FSD}(\bar{r}) = \texttt{Agree}(\bar{r}_1) - \texttt{Agree}(\bar{r}_2)
\]
\(\bullet\) \emph{Answer-Number-based}~\citep{DBLP:journals/corr/abs-2403-09849}:\\
Again we derive a set of unique answers from the model’s output, denoted as \(\mathbf{\hat{r}}\). The answer-number-based , \(\texttt{Ans-Num}(\bar{r})\), is then simply defined as:
\[
\texttt{Ans-Num}(\bar{r})=1-\frac{|\mathbf{\hat{r}}|}{n}
\]
\(\bullet\) \emph{Pairwise-Comparison-based}~\citep{DBLP:journals/corr/abs-2403-09849}:\\
We count the number of times each different answer is repeated, denoted as \(n_{1}\ldots n_{|\mathbf{\hat{r}}|}\). Specially, we denote the index of the majority voting result as \(i_{\bar{r}}\). Then the pairwise-comparison-based, \(\texttt{Pairwise}(\bar{r})\), metric is defined as:
\[
\texttt{Pairwise}(\bar{r})=\prod^{|\mathbf{\hat{r}}|}_{j \neq i_{\bar{r}}} \frac{n_{i_{\bar{r}}}}{n_{i_{\bar{r}}}+n_j}
\]
We evaluate the performance of Mirror-Consistency and Self-Consistency on the task of confidence calibration using the five distinct metrics described above. Our experiments span two datasets, GSM8K and SVAMP, and consider two models: GPT-3.5-turbo and Qwen-turbo. 

\begin{figure*}[t!]
\centering
\includegraphics[width=\textwidth]{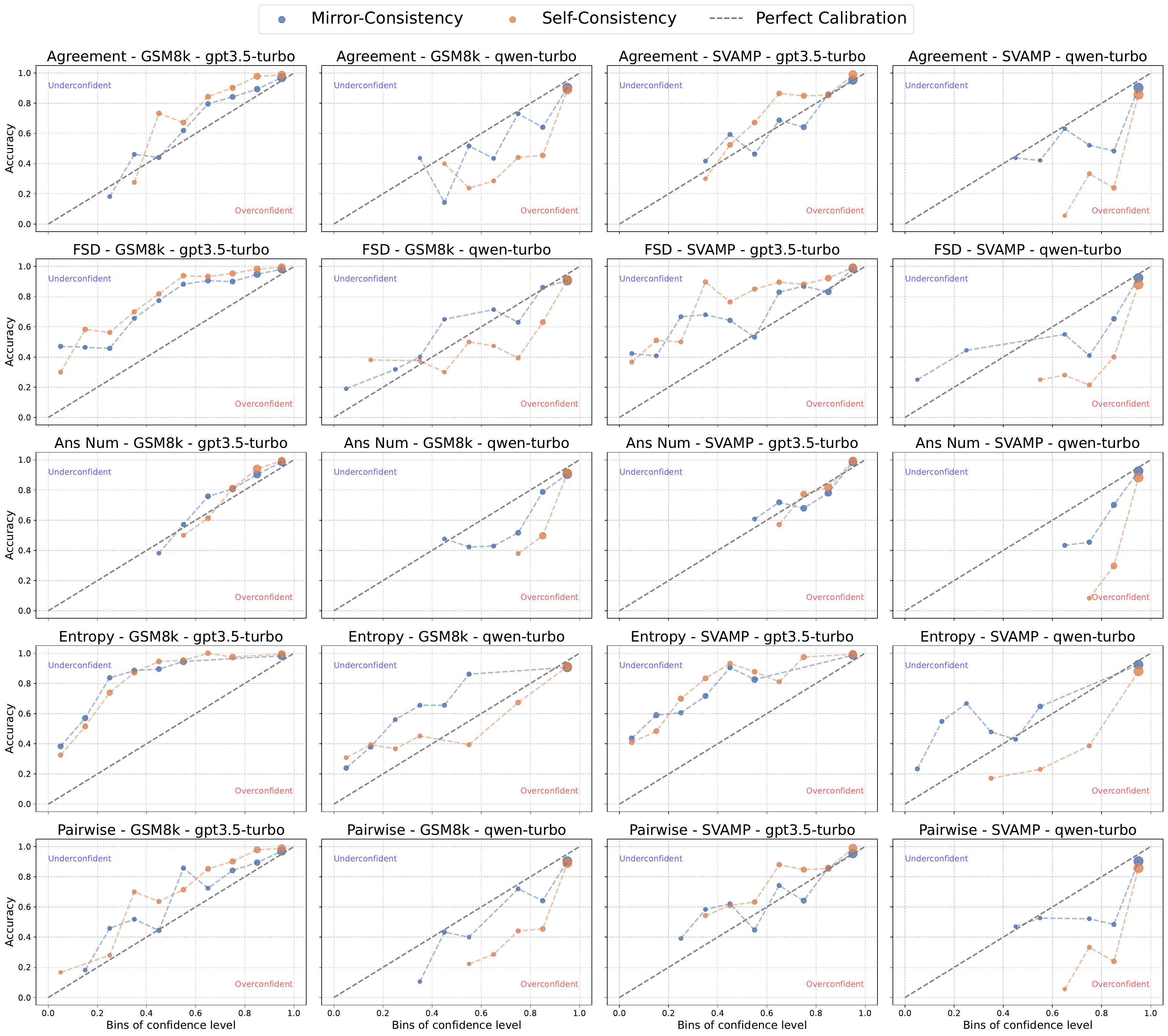}
\caption{Visualization of confidence calibration results across different calibration metrics. We set the number of bins as 10. We omit the bins if their size is less than 1\% of the total dataset. The point size is proportional to the square root of its corresponding bin size.}
\vspace{-0.13in}
\label{fig: full calibration curves}
\end{figure*}

\paragraph{Main result} As shown in \autoref{tab: Full ECE Scores Comparison}, we find that Mirror-Consistency outperforms Self-Consistency most of the time. Additionally, we provide the corresponding calibration curves in \autoref{fig: full calibration curves}. We set the number of bins as 10. We do not plot bins with a bin size smaller than 1\% of the total data set. The size of the points is proportional to (the square root) of the bin size. We find that in most cases Mirror-Consistency's calibration curve is closer to the ideal curve. It is noteworthy that when methods based on Self-Consistency encounter issues of overconfidence (where the calibration curve tends toward the bottom right), Mirror-Consistency can significantly mitigate the phenomenon of overconfidence. This indicates that multiple comparisons and reflections against minority opinions can alleviate the overconfidence that arises from solely relying on majority opinions.

\section{Related Works}

\subsection{Self-Correction in Languages Models}
Recent advancements in large language models have shed light on their advanced cognitive intelligence, notably their ability for self-correction. This attribute enables LLMs to amend their initial responses by integrating both external and self-generated feedback, thus improving previous outputs~\citep{ Xi2023SelfPolishER, Pan2023AutomaticallyCL, Nathani2023MAFMF}. The concept of self-correction encompasses a variety of techniques that have been extensively surveyed and categorized based on the source of feedback and the timing of correction~\citep{Pan2023AutomaticallyCL}. These techniques range from using explicit error messages in tasks, such as code execution, to human feedback and model-generated prompts~\citep{ miao2024selfcheck,  chen2023iterative, DBLP:journals/corr/abs-2310-01798, kim2023language}. While iterative prompting techniques have shown promise, recent research has raised concerns about LLMs' capacity for independent reflection, revealing limitations in modifying responses without external feedback~\citep{DBLP:journals/corr/abs-2310-01798, Stechly2023GPT4DK, Liang2023EncouragingDT, DBLP:journals/corr/abs-2311-08516}.

\subsection{Confidence Calibration}
In the realm of machine learning, uncertainty quantification methods are crucial for assessing the risk associated with model predictions~\citep{ruping2006robust,desai-durrett-2020-calibration}. Traditional calibration approaches, including probabilistic, ensemble-based, and density-based methods, although effective, require extensive computational resources and access to model internals, making them less viable for closed-source LLMs \citep{calibration-guo, lakshminarayanan_simple_2017, gal_dropout_2016, lee_simple_2018, yoo_detection_2022}. Furthermore, some post-hoc strategies have emerged, focusing on eliciting model-estimated probabilities of correctness or directly verbalized confidence levels \citep{kadavath2022language, lin2022teaching, mielke-etal-2022-reducing}. And recently, inspired by the sample-based methods, calibration through sample consistency, which relies solely on model input and output, has been explored. \citep{wang2023selfconsistency, selfcheckgpt, DBLP:journals/corr/abs-2306-13063, portillo-wightman-etal-2023-strength}.

\subsection{Refined Problem-Solving Strategies}
Existing prompting strategies engage in two forms of reasoning: exploring various perspectives (breadth) and refining ideas to reduce errors (depth). Self-consistency and related methods~\citep{Wang2023Self, Huang2022LargeLM, Yoran2023AnsweringQB, Jain2023SelfconsistencyFO} promoting breadth through diverse reasoning sampling. In contrast, strategies like self-reflection and abstraction~\citep{Shinn2023ReflexionAA, Madaan2023SelfRefineIR, Paul2023REFINERRF, Zheng2023ProgressiveHintPI} focus on the depth of reasoning by iteratively refining prompts.

\section{More Implementation Details}
\subsection{Dataset Details}
\paragraph{Math Word Problems (MWP).} This category includes challenges where the objective is to compute numeric solutions to problems framed in natural language. Our analysis incorporates several datasets, each selected to evaluate distinct aspects of mathematical reasoning. For \textbf{GSM8K} \cite{cobbe_training_2021}, participants engage with a diverse set of elementary math questions, testing their ability to apply basic arithmetic operations and logical reasoning. Meanwhile, \textbf{SVAMP} \cite{patel_are_2021} is specifically designed to probe the robustness of models against changes in question phrasing and structural complexity, offering a stringent test of comprehension and adaptability.

\paragraph{Multi-hop QA.} Tasks in this section demand answers to intricate questions via a series of logical deductions, necessitating a nuanced understanding of the content. Answers may be in the form of Boolean values or specific textual responses. \textbf{StrategyQA} \cite{geva_did_2021} dataset presents science questions that require an implicit strategy for multi-step reasoning, challenging the model's ability to form and execute complex inferential chains. \textbf{Date Understanding} \cite{big-bench_collaboration_beyond_2021} tests temporal reasoning by asking participants to calculate dates based on relative time information, thus assessing both numerical and chronological reasoning skills.

\subsection{Model Details}\label{sec: Model Details}
\paragraph{Closed-Source Language Models}
We utilize two advanced closed-source language models, GPT-3.5-turbo and Qwen-turbo. For GPT-3.5-turbo, the API is accessible at \url{platform.openai.com}. Qwen-turbo, another potent large language model, is available via the DashScope provided by Alibaba Cloud (\url{alibabacloud.com}). Detailed guidance about DashScope and its integration can be accessed \href{https://github.com/QwenLM/Qwen}{here}. Further information about utilizing Qwen-turbo through Alibaba Cloud, including API instructions, can be found \href{https://alibabacloud.com/blog/generative-ai-with-function-compute_601032}{here}.
\paragraph{Open-Source LMs} We use the newly released Llama-3 Family (\url{llama.meta.com}).  We conduct experiments using Nvidia A100 80GB GPUs. Specifically, \textit{Meta-Llama-3-70B-Instruct} requires two GPUs per inference, whereas \textit{Meta-Llama-3-8B-Instruct} operates with a single GPU. 
\subsection{Other Details}
\paragraph{Evaluation Metrics} We use Expected Calibration Error (ECE)~\citep{calibration-guo} as the calibration metric. Predictions are first binned into \(M = 10\) intervals based on model confidence. For each bin \(B_m\), we calculate the average accuracy \(\texttt{acc}(B_m)\) and average confidence \(\texttt{conf}(B_m)\). ECE is defined as their weighted absolute differences:
\[
\texttt{ECE} = \sum_{m=1}^M \frac{|B_m|}{N} \left|\texttt{acc}(B_m) - \texttt{conf}(B_m)\right|
\]

\paragraph{Split Setting} In order to save computational resources without loss of generality, we select the first 768 examples from the test split of GSM8K and SVAMP for our experiments. We use the full split of Date. For StrategyQA, we test on the first 490 samples following \citet{shridhar2023screws}.


\section{Prompt Template}\label{sec:prompt}
In this section, we outline the three crucial prompts required for the future realization of Mirror-Consistency: 1) Prompt for Simple Resampling, denoted as $p_{\text{sample}}$, 2) Prompt for Reflection on Inconsistency, denoted as $p_{\text{contrast}}$, 3) Prompt for Conditional Resampling, denoted as $p_{\text{sample w/ fb}}$.

It is important to note that our implementation adopts the Chain-of-Thought (COT) approach to resample multiple answers. However, it is feasible to explore a variety of other prompt methods, such as Least-to-Most (LtM) ~\citep{zhou2023leasttomost}, Program of Thoughts (PoT)~\citep{chen_close_2023} and Faithful CoT (FCoT)~\cite{fcot}. Furthermore, in our experiments to validate the generality of Mirror-Consistency, we apply the same prompts across diverse datasets, including arithmetic, symbolic reasoning, and commonsense reasoning. Nonetheless, future work may tailor prompts specifically to the characteristics of different datasets, thus enhancing reflection and contrast, to potentially improve Mirror-Consistency.

\subsection{Prompt for Simple Resampling $p_{\text{sample}}$}
The following template is employed in a simple resampling process. During its use, \textbf{[QUESTION]} must be substituted with the specific question at hand. This template is designed for initial sampling in Mirror-Consistency and for instances when the Checklist is empty, as well as serving as a baseline in our experiments with Self-Consistency.
\begin{tcolorbox}[colback=gray!5!white,colframe=gray!75!black,title=Simple Resampling Prompt]
Solve the following problem step by step. Begin each step with "Step :" and ensure each step is separated by "\textbackslash n\textbackslash n". Conclude with the phrase "So the answer is", followed by the answer.\\

Question: \textbf{[QUESTION]}

\noindent\rule{\textwidth}{0.4pt}
Answer:
\end{tcolorbox}
\subsection{Prompt for Contrast $p_{\text{contrast}}$}
The template for reflecting on inconsistencies is as follows. In practice, \textbf{[QUESTION]} should be replaced with the actual question, \textbf{[PRE-MAJORITY-VOTE]} should be replaced with the majority vote answer from all previous rounds (in cases of multiple identical responses, one is randomly selected as the majority vote answer), \textbf{[CUR-RESPONSE]} should be substituted with the most recently generated response of the current round, and finally \textbf{[PRE-CHECKLIST]} should be replaced by the checklist from the last round. This template facilitates the reflection of inconsistencies within the Mirror-Consistency approach.

\begin{tcolorbox}[colback=red!5!white,colframe=red!75!black,title=Reflection on Inconsistency Prompt]
Given two candidate solutions for a question, carefully analyze and compare the differences in their reasoning steps. Consider: 1) The specific differences in their reasoning steps and final answers; 2) The reasons behind these disparities.\\

Question: \textbf{[QUESTION]}\\

Two solutions: \\
Solution 1: \textbf{[PRE-MAJORITY-VOTE]}\\
Solution 2: \textbf{[CUR-RESPONSE]}\\

If no differences exist, output \textless STOP!\textgreater.\\
If differences are identified, describe them, determine errors, and explain why. Extract a key suggestion to prevent such errors and combine it with the previous checklist \textbf{[PRE-CHECKLIST]} to formulate a new checklist. Begin the checklist with \textless CHECKING\textgreater.

\noindent\rule{\textwidth}{0.4pt}
Feedback:
\end{tcolorbox}

\subsection{Prompt for Conditional Resampling $p_{\text{sample w/ fb}}$}
In the conditional resampling process, the following template is utilized. It requires replacing \textbf{[QUESTION]} with the specific question and substituting \textbf{[CHECKLIST]} with the latest checklist derived from the previous contrast phase. This template is used in Mirror-Consistency for resampling new responses based on reflective feedback addressing inconsistencies.
\begin{tcolorbox}[colback=blue!5!white,colframe=blue!75!black,title=Conditional Resampling Prompt]
Solve the following problem step by step. Begin each step with "Step :" and ensure each step is separated by "\textbackslash n\textbackslash n". Conclude with the phrase "So the answer is", followed by the answer.\\

Consider integrating the previous advice: \textbf{[CHECKLIST]}, into your solution process.\\

Question: \textbf{[QUESTION]}

\noindent\rule{\textwidth}{0.4pt}
Answer:
\end{tcolorbox}

\begin{table*}
\footnotesize 
\centering
\small
\label{tab:datasets}
\begin{tabularx}{\textwidth}{@{}lXXX@{}}
\toprule
\textbf{Dataset} & \textbf{Reference} & \textbf{URL} & \textbf{License} \\
\midrule
\textbf{Math Word Problems} & & & \\
GSM8K & Cobbe et al. (2021) \cite{cobbe_training_2021} & \url{https://github.com/openai/grade-school-math} & MIT License: \url{https://github.com/openai/grade-school-math/blob/master/LICENSE} \\
SVAMP & Patel et al. (2021) \cite{patel_are_2021} & \url{https://github.com/arkilpatel/SVAMP} & MIT License: \url{https://github.com/arkilpatel/SVAMP/blob/main/LICENSE} \\
\midrule
\textbf{Multi-hop QA} & & & \\
StrategyQA & Geva et al. (2021) \cite{geva_did_2021} & \url{https://github.com/google/BIG-bench/tree/main/bigbench/benchmark_tasks/strategyqa} & Apache License v.2: \url{https://github.com/google/BIG-bench/blob/main/LICENSE}\\
Date Understanding & BIG-Bench Collaboration (2021) \cite{big-bench_collaboration_beyond_2021} & \url{https://github.com/google/BIG-bench} & Apache License v.2: \url{https://github.com/google/BIG-bench/blob/main/LICENSE} \\
\bottomrule
\end{tabularx}
\caption{Summary of URLs and Licenses}
\label{tab: Summary of URLs and Licenses}
\end{table*}
\section{URLs and Licenses}
\autoref{tab: Summary of URLs and Licenses} provides license information for the datasets we utilize in our experiments. We employ all the above datasets solely for research purposes, in accordance with their designated uses.

\end{document}